\documentclass[journal]{ijiss}
\usepackage{graphicx}
\usepackage{url,cite}
\usepackage{booktabs}
\usepackage{subfigure,array,rotating,amsmath}
\usepackage[table]{xcolor}
\usepackage{multirow}

\ifCLASSINFOpdf
\else
\fi

\begin{document}

%
\title{3D Reconstruction of Crime Scenes and Design Considerations for an Interactive Investigation Tool}

\author{Erkan~Bostanci
\address{Ankara University, Faculty of Engineering, Computer Engineering Department.\protect\\
Ankara University, Golbasi 50.Yil Yerleskesi, Bahcelievler Mah. 06830, Ankara, Turkey. Tel: +90-203-3300/1767.
e-mail: ebostanci@ankara.edu.tr}}

\ijisstitleabstractindextext{%
\begin{abstract}
Crime Scene Investigation (CSI) is a carefully planned systematic process with the purpose of acquiring physical evidences to shed light upon the physical reality of the crime and eventually detect the identity of the criminal. Capturing images and videos of the crime scene is an important part of this process in order to conduct a deeper analysis on the digital evidence for possible hints. This work brings this idea further to use the acquired footage for generating a 3D model of the crime scene. Results show that realistic reconstructions can be obtained using sophisticated computer vision techniques. The paper also discusses a number of important design considerations describing key features that should be present in a powerful interactive CSI analysis tool.

\end{abstract}

\begin{ijisskeywords}
Crime scene investigation, forensic science, 3D reconstruction, computer vision, design.
\end{ijisskeywords}}

\maketitle



%

\section{Introduction}

The environment where a crime was committed, a crime scene, can be considered as the natural witness of the crime. This silent witness, indeed, proves to be very helpful in finding crucial evidences related to the story of the crime and allows investigators to establish the correct reasoning and hence to gain a valuable insight into how the crime was committed~\cite{Ccb2013,Criminalistics1993}. 

It is known that the crime scene must be investigated using a detailed systematic approach~\cite{Girard2008, Forest1983, Saferstein1998}. The stages of Crime Scene Investigation (CSI) can be summarized as follows: First stage is receiving information about a crime, planning/preparation for the investigation and eventually arriving at the crime scene. Upon arrival, a preliminary investigation is performed for deciding the appropriate equipment and material for collecting the evidence. 

\begin{figure}[h!t!p!b]
	\begin{center}
		\includegraphics[width=\columnwidth]{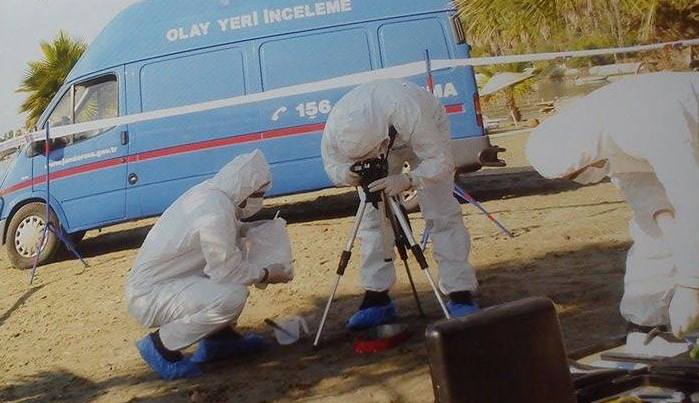}
	\end{center}
	\caption{A Gendarmarie CSI team investigating a crime scene. Figure courtesy of Gendarmarie Schools Command.}
	\label{fig:csi}
\end{figure}

In the meantime, wounded or injured people and casualties are identified after doctor's medical examination and wounded or injured people are transferred to the hospital. The next stage is capturing images or videos of the crime scene~\cite{Csi2001} and preparing a sketch diagram of the environment which also indicates the camera's positions while the images are being captured as shown in Fig.~\ref{fig:csi}. The evidences found in the scene \emph{e.g.} fingerprints, footprints, pieces of clothing, hair, \emph{etc.} are also shown in this sketch using the identification numbers assigned to them. The distances between these evidences and the victim's position are also annotated in the sketch. Following stages comprise the collection and packaging of the physical evidences in the environment. Finally, the evidences are transferred to the forensic laboratory for detailed analysis. All these stages (and several sub-stages which were not mentioned here) constitute and require a meticulous process for a successful investigation~\cite{Cic2008}.

\begin{figure}[h!t]
	\begin{center}   
		\subfigure[]{\includegraphics[width=\columnwidth]{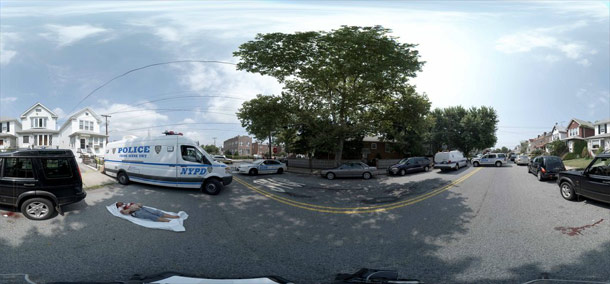}}
		\subfigure[]{\includegraphics[width=0.45\columnwidth]{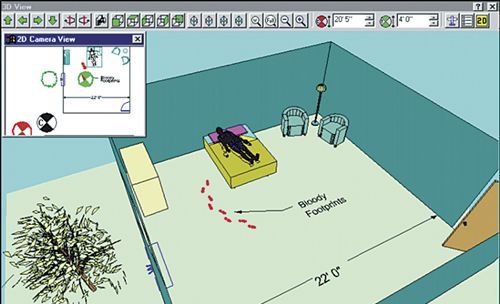}}
		\subfigure[]{\includegraphics[width=0.45\columnwidth]{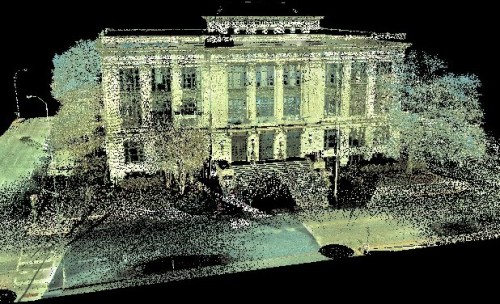}}
	\end{center}
	\caption{Different approaches used for modelling a crime scene. (a) Using panoramic images. (b) Sketches and 3D modelling. (c) Laser scanner generated point clouds. Figures courtesy of~\cite{Panorama2014,Sketch2014,Laser2014} respectively.}
	\label{fig:methods}
\end{figure}

The work presented here is related to the stage where images or videos of the crime scene are captured. Conventionally, the crime scene is captured by taking a number of photos using several different viewpoints as well as capturing images of individual items that may prove to be useful evidence. Panoramic images, see Fig.~\ref{fig:methods}(a), also allow viewing the crime scene from a wider angle. In order to show the spatial structure of the environment in 3D, rather than relying solely on images or sketches, an approach is to create a 3D model of the crime scene using modelling tools such as 3D Max. There are also companies providing tailored software solutions for this purpose (\emph{e.g.}~\cite{Software2014}) facilitating the 3D modelling process for crime scenes as shown in Fig.~\ref{fig:methods}(b). More recent approaches make use of laser scanners to reconstruct the crime scene in form of a point cloud (Fig.~\ref{fig:methods}(c)) which is very useful for representing the 3D structure of the environment~\cite{Bostanci2013, Bostanci2012b}.

Capturing images of the complete scene can provide useful evidence; however, the images can be misleading due to the very nature of the imaging process itself (perspective projection, lens distortions, \emph{etc}.~\cite{Bradski2008}) Panoramic images, likewise, are affected by similar problems. Manual modelling of the scene is laborious and may not result in a realistic model; though addition of annotations of distances between objects can be quite useful. 3D laser scanners~\cite{Faro2014}, a technique known as structured light, produce accurate point clouds up to a range of 120m~\cite{FaroReview2014}. The problem with these is that laser beams projected from the device may not always reflect back to the sensor properly due to transparent surfaces (\emph{e.g.} windows) or corners. Concerns related to the pricing of such devices, which is in the order of ten thousand dollars, are additional parameters to take into account.

Proposed approach captures the crime scene in form of a video sequence and then extracts frames, indeed keyframes, from this sequence. The keyframes (\emph{i.e.} systematically selected frames from a sequence) are used to reconstruct the scene in 3D using the dense reconstruction approach of \cite{Snavely2006, Furukawa2010} which employs an iterative technique known as bundle adjustment (see Section~\ref{sec:denseReconstruction})~\cite{Triggs2000, Szeliski2011}. This reconstruction is basicly a list of 3D points which is later used by other third party software packages for editing the point cloud resulting in a metric, realistic model of the crime scene. 

The rest of the paper is structured as follows: Section~\ref{sec:acquisition} describes the approach for deciding which frames from the complete sequence will be used for reconstruction. Section~\ref{sec:denseReconstruction} is dedicated for the dense reconstruction of the crime scene from selected frames followed by Section~\ref{sec:postProcessing} where the point cloud is edited for details. Results are presented in Section~\ref{sec:results}, and then Section~\ref{sec:design} presents crucial system requirements for developing an interactive analysis tool that will be useful throughout the investigation process. Finally, the paper is concluded in Section~\ref{sec:conclusion}.

\section{Acquiring the Video Sequence and Keyframing}
\label{sec:acquisition}

When performing 3D reconstruction on a video sequence, it is useful to select a representative subset of the frames, a process known as keyframing. In this process, it is important to select these keyframes so that the number of feature correspondences between them is above some threshold. This allows the system to estimate the 3D structure correctly and not skip any important image information. Keyframing offers the following two benefits~\cite{Bostanci2012c}:

\begin{itemize}
	\item Large baseline: The estimation of 3D structure is not accurate when the baseline is small \emph{i.e.} there is not enough motion (\emph{e.g.} translation or rotation) between the keyframes.
	
	\item Performance: The complete algorithm is run on a smaller set of images instead of the complete sequence, hence the time spent for reconstruction will be reduced.
\end{itemize}

The keyframing approach described in the following paragraph is based on~\cite{Mouragnon2006, Royer2005}. The work described in~\cite{Bostanci2012c} used an extra heuristic to prevent skipping too many frames as this may result in missing any sudden motion that may have lasted for just a few frames. However, this approach was not employed in this work since the environment is still and camera is assumed to have a slow and smooth motion.

\begin{figure*} [h!t!p]
	\begin{center}
		\includegraphics[width=0.8\textwidth]{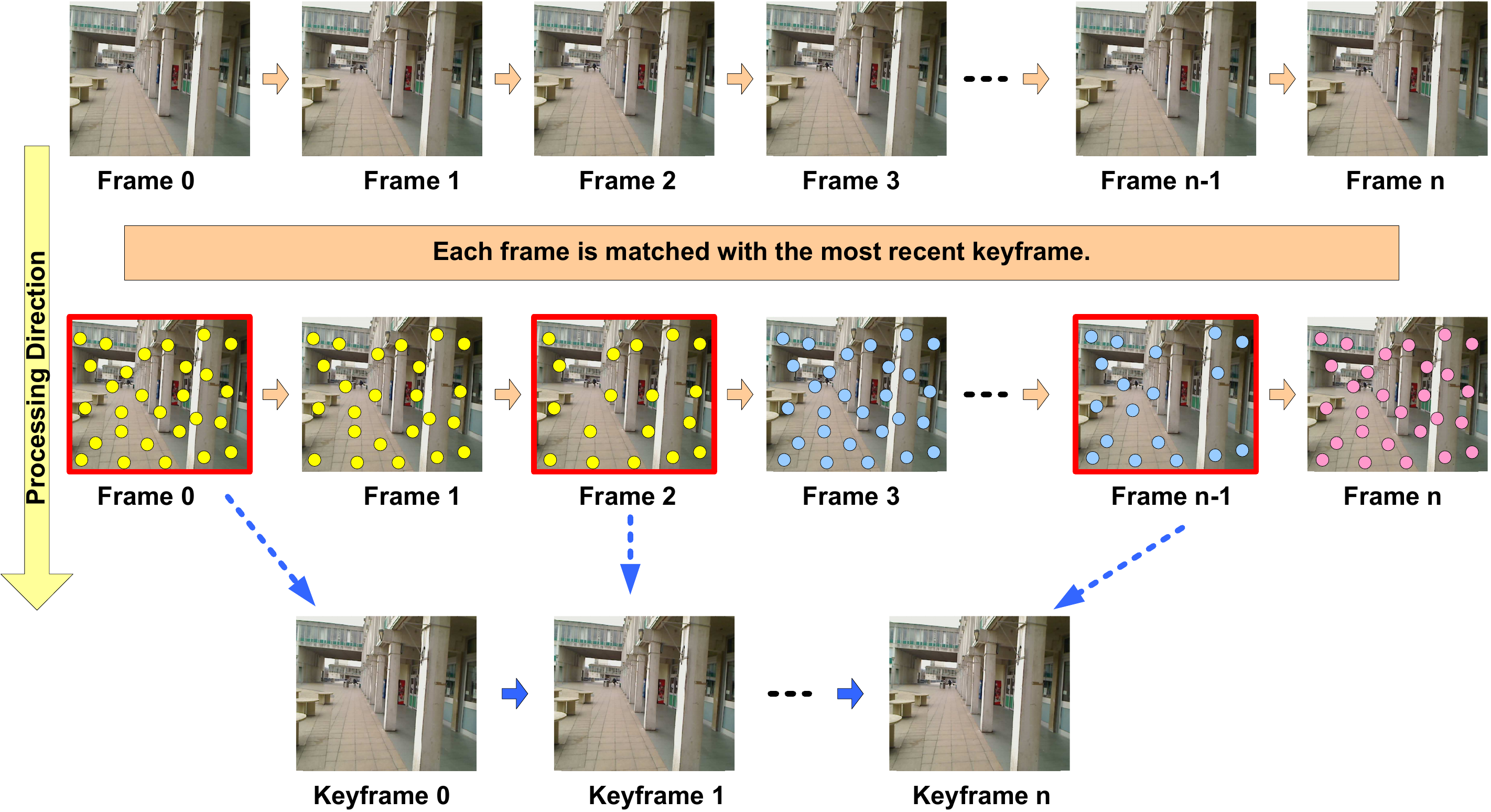}
	\end{center}
	\caption{Selecting keyframes based on image correspondences. Coloured circles represent feature correspondences between frames acquired by the camera and the most recent keyframe. When the number of these correspondences fall below a threshold, a new keyframe is extracted.}
	\label{fig:keyframes}
\end{figure*}

The first image acquired by the camera, $frame_0$, is always selected as the first keyframe ($keyframe_0$). For the following frames, a set of feature correspondences were computed between the last keyframe $keyframe_n$ and each of following frames $frame_i$. Spurious matches are eliminated from this set using RANSAC~\cite{Fischler1981}. When the number of these correspondences (\emph{i.e.} inliers) falls below a threshold ($t=200$, found experimentally), a new keyframe is extracted as $keyframe_{n+1}$. This selection mechanism is illustrated in Fig.~\ref{fig:keyframes}.

It is important to note that repeatability~\cite{Schmid2000} of a feature detector is an important concern in this approach for keyframe extraction. Repeatedly detected and matched features in consecutive frames suggest that common features are available in the environment and are captured in these frames. For this reason, successful matches between different video frames captured with significant amount of time eliminate the redundancy in extracted keyframes.

The keyframing approach used \emph{SIFT}~\cite{Lowe2004} and \emph{SURF}~\cite{Bay2008} detector/descriptors for their several advantages in detecting scale and affine invariant features, good matching performance and near-to-uniform feature coverage across the image~\cite{Bostanci2013b}.

This method of extracting keyframes uses $24\%$ -- $89\%$ of the frames depending on the test sequence --since several parameters such as motion blur and noise can affect features correspondences-- and provides a large enough baseline for accurate 3D reconstruction as well as allowing better speed performance. 

\section{Dense Reconstruction}
\label{sec:denseReconstruction}

In order to create a reconstruction of the environment, one first needs to find the camera poses from which the images have been captured. After finding the camera poses, the next stage is triangulation for calculating the 3D information for the image features used for finding the camera pose. In practice, these two stages are very dependent on each other. In other words, one needs to find the correct solution of the camera matrix out of four possible solutions by triangulating the features for each solution and choosing the one with the highest number of 3D inliers after projection~\cite{Hartley2003}. An initial pose estimate can be obtained in this way.

Due to several parameters, \emph{e.g.} noise, lens distortions, \emph{etc.}, that are very natural to the imaging process, this initial pose estimate may not be very accurate and hence must be refined. Bundle adjustment~\cite{Triggs2000} is an iterative method for refining the camera pose estimates and 3D point coordinates using an optimization technique (usually Levenberg–-Marquardt~\cite{Kelley1999}). The aim is to minimize the projection errors for the 3D points using the camera parameters that model its position and orientation. This approach is computationally expensive and, even for a small number of camera parameters, may take hours to converge. Sparse versions of the algorithm, aimed at improved efficiency, are also available~\cite{Lourakis2009}.

The software package \emph{Bundler} developed by Snavely et al.~\cite{Snavely2006} allows finding camera parameters described above and computing \emph{sparse} 3D structure of the image features extracted from a set of images. The package was initially developed for an unordered collection of images; however, it displays similar performance for ordered sets of images such as the keyframes from a video sequence used in this work. It is worth mentioning that \emph{Bundler} is based on an assumption that there is enough disparity \emph{i.e.} images taken from viewpoints those are well apart from each other. This is hard to satisfy in a video sequence where there will be many frames of the same scene from the same viewpoint. The keyframing approach described proved to be very useful in such a scenario.

\begin{figure} [h!t!p]
	\begin{center}
		\includegraphics[width=\columnwidth]{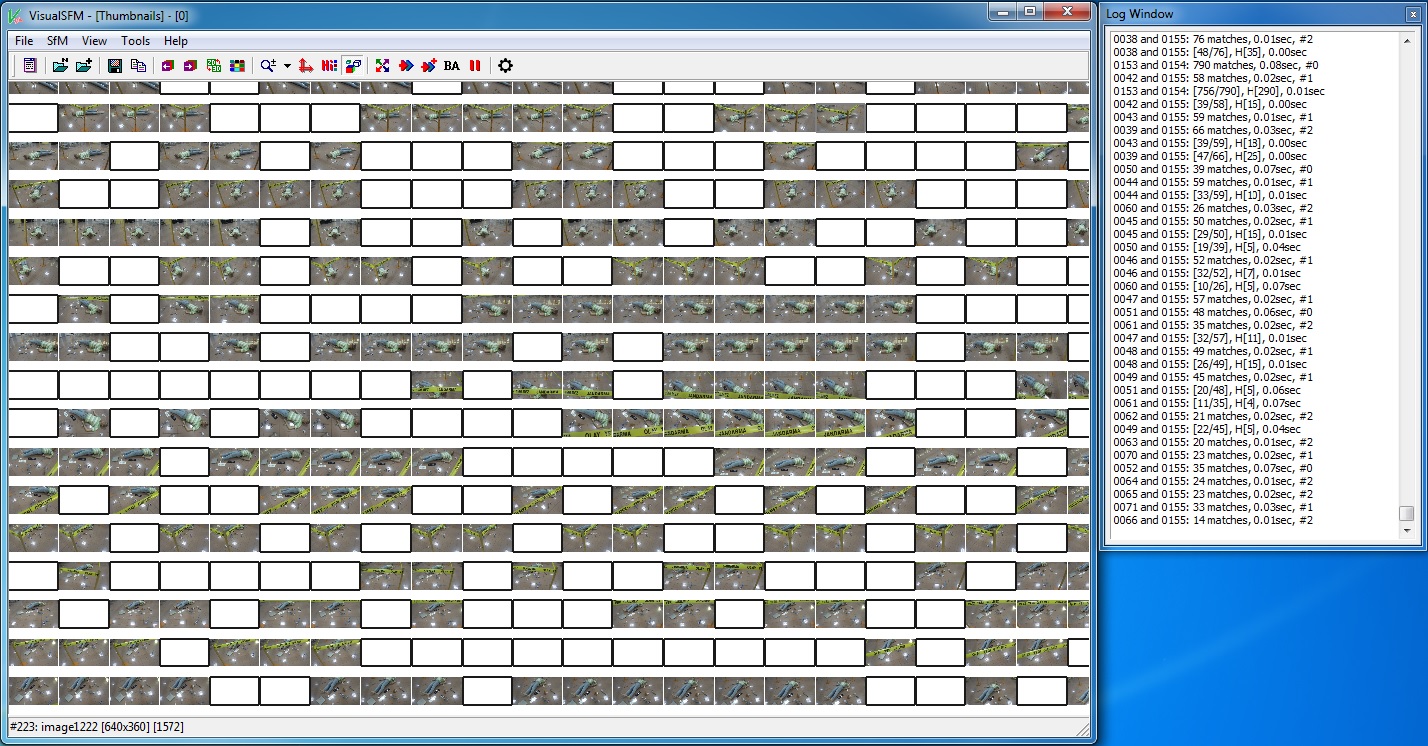}
	\end{center}
	\caption{Finding the camera parameters. The software displays thumbnails of the extracted keyframes and matching is performed on the right.}
	\label{fig:sfm}
\end{figure}

For dense reconstruction, a different library called \emph{CMVS} (Clustering Views for Multi-view Stereo) by Furukawa and Ponce~\cite{Furukawa2010} was employed. \emph{CMVS} takes the set of images and the corresponding camera parameters, found using \emph{Bundler}, and then performs \emph{dense} reconstruction of the scene by clustering the complete input image set into smaller sets. The output is obtained in form of several clusters of 3D dense point clouds.

\section{Post Processing}
\label{sec:postProcessing}

Obtained set of point clouds correspond to different clusters from the output of \emph{CMVS} and contain valuable information for the crime scene and hence selecting one point cloud and leaving others cannot be an option. 

One problem with these point clouds is that they may not be aligned to each other. This is where some manual processing is required. Fortunately, there are tools that facilitate this process such as \emph{MeshLab}~\cite{MeshLab2014} and \emph{CloudCompare}~\cite{CloudCompare2014}.

These tools employ a semi-automatic way for aligning two misaligned point clouds, one is called \emph{model} (fixed one) and the other is \emph{data} (one to be aligned), using the \emph{ICP} (Iterative Closest Point) algorithm~\cite{Besl1992}. The algorithm required at least 4 point correspondences from the two point clouds. Once these 4 corresponding points are selected manually by the user, the software calculates the required transformation for the \emph{data} to map it onto the \emph{model}. Considering the errors made by the user during selection of the corresponding points, the algorithm computes a rough transformation to align the two point clouds.

Following this mapping, a global alignment stage is performed in order to reduce the error in the alignment. This is again an iterative process, constantly reducing the error bound.

One thing to note, the statistical outlier removal tool available in \emph{CloudCompare} proved very useful for deleting the orphaned points or noise from the point cloud prior to alignment in \emph{MeshLab}.

\section{Results}
\label{sec:results}

\begin{figure}[h!t]
	\begin{center}   
		\subfigure[]{\includegraphics[width=0.45\columnwidth]{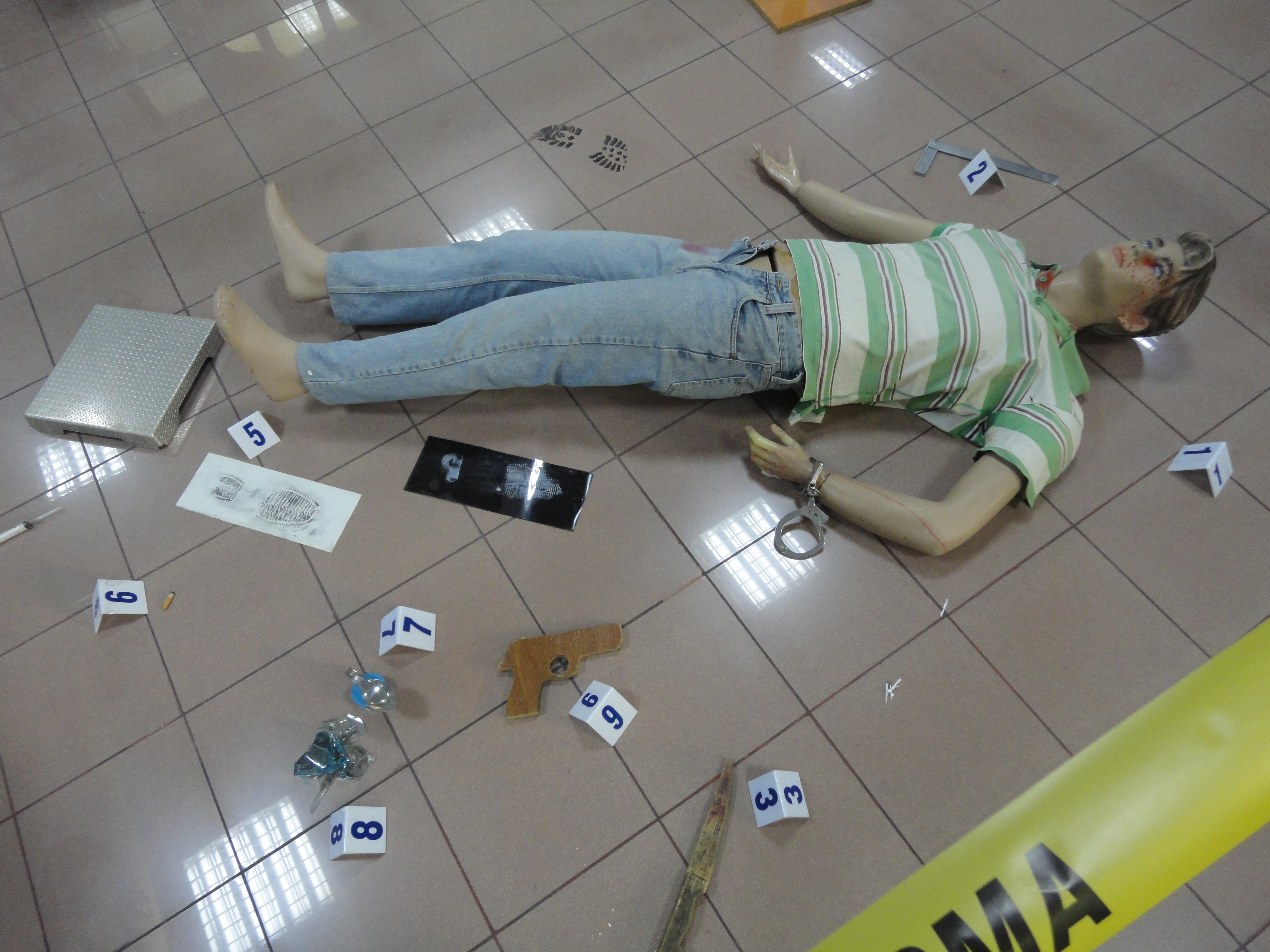}}
		\subfigure[]{\includegraphics[width=0.45\columnwidth]{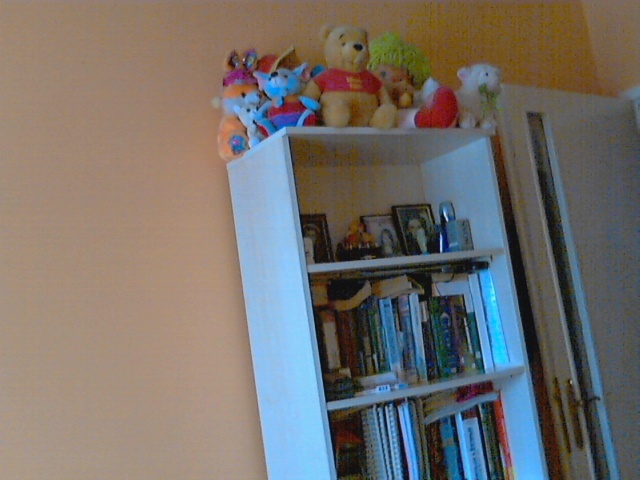}}
	\end{center}
	\caption{Test environments.}
	\label{fig:sequences}
\end{figure}

Test sets used for the experiments were captured in different environments: first two sets (resampled to 640$\times$360 pixels from an initial size of 1280$\times$720 pixels) were captured in a CSI training lab (Fig.~\ref{fig:sequences}(a)) and the last set was captured a small room environment as shown in Fig.~\ref{fig:sequences}(b). The last one was captured using a simple webcam hence motion blur is quite noticeable in this test sequence.

The detector/descriptor parameters were used as the default ones recommended by their authors and they are as follows: 128 bin descriptors, 4 octaves, 5 and 2 octave layers for \emph{SIFT} and \emph{SURF} respectively.

Table~\ref{tab:keyframeNumbers} shows the number of total frames in the test sequences and the extracted keyframes when \emph{SIFT} and \emph{SURF} detector/descriptors are used. The fewer number of the extracted keyframes for \emph{SIFT} is suspected to be due to its superiority in terms of repeatability (see~\cite{Schmid2000} for details) over that of \emph{SURF}, though a more detailed statistical analysis is required to be more conclusive about this result. When a feature is detected repeatedly and matched successfully, it is added to the number of inliers, which is the main criterion for selecting keyframes as described in Section~\ref{sec:acquisition}. 

\begin{table}[htbp]
	\centering
	\caption{Number of extracted keyframes for \emph{SIFT} and \emph{SURF}}
	\begin{tabular}{cccc}
		\toprule
		\multirow{2}[0]{*}{\textbf{Test Sequence }} & \multirow{2}[0]{*}{\textbf{\# of Frames}} & \multicolumn{2}{c}{\textbf{\# of Extracted Keyframes}} \\
		&       & \emph{SIFT}  & \emph{SURF} \\
		\textbf{1} & 3675  & 867 (23.59\%) & 1012 (27.54\%) \\
		\textbf{2} & 3645  & 1339 (36.74\%) & 1519 (41.67\%) \\
		\textbf{3} & 964   & 853 (88.49\%) & 755 (78.32\%) \\
		\bottomrule
	\end{tabular}%
	\label{tab:keyframeNumbers}%
\end{table}%

The advantage of having fewer keyframes is that the bundle adjustment process will be completed sooner. It is also important to reiterate that these keyframes must be representative of the entire sequence. The noticeable motion blur in the third test sequence resulted in a larger number of keyframes for both detectors. Needless to say, \emph{SURF} showed its speed advantage in the matching process for keyframe extraction, though resulted in a larger number of keyframes.

\begin{figure} [h!t!p]
	\begin{center}
		\includegraphics[width=0.8\columnwidth]{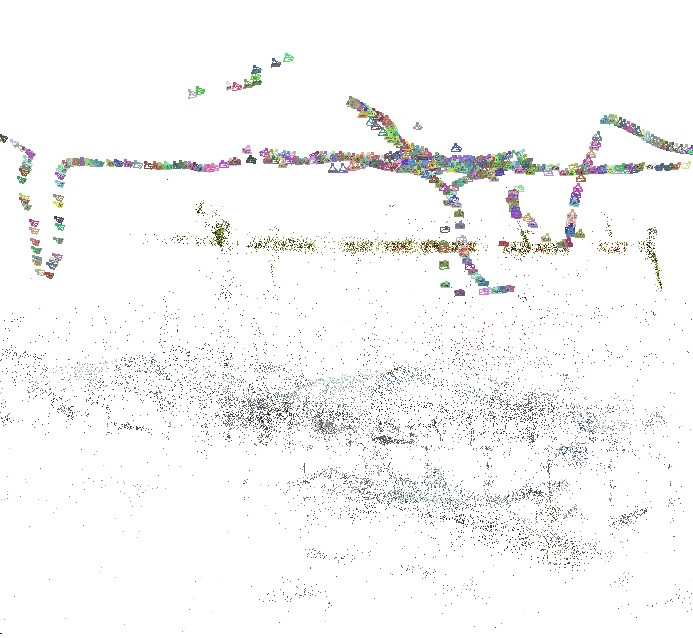}
	\end{center}
	\caption{Reconstructed camera path and the 3D point cloud of the crime scene.}
	\label{fig:reconstruction}
\end{figure}

Using the keyframes extracted, a sparse reconstruction of the scene along with the camera path is obtained as shown in   Fig.~\ref{fig:reconstruction}.

\begin{figure} [h!t!p]
	\begin{center}
		\includegraphics[width=0.8\columnwidth]{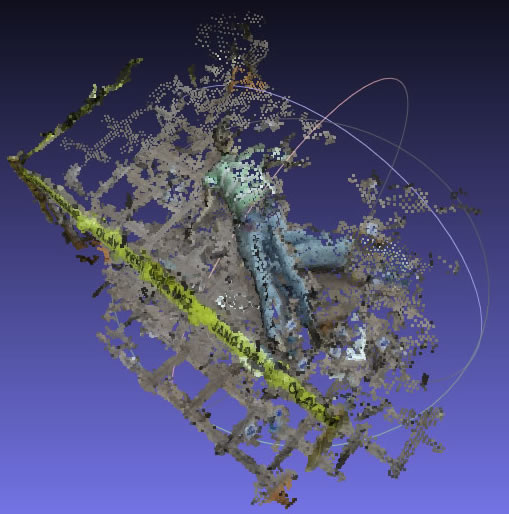}
	\end{center}
	\caption{Two or more point clouds can be misaligned after dense reconstruction.}
	\label{fig:misliagned}
\end{figure}

Section~\ref{sec:postProcessing} mentioned that there could be alignment problems between various point clouds obtained from the same scene since \emph{CMVS} works by dividing the complete scene into image clusters first and then performs dense reconstruction. Fig.~\ref{fig:misliagned} depicts the misaligned clouds superimposed together.

\begin{figure} [h!t!p]
	\begin{center}
		\includegraphics[width=\columnwidth]{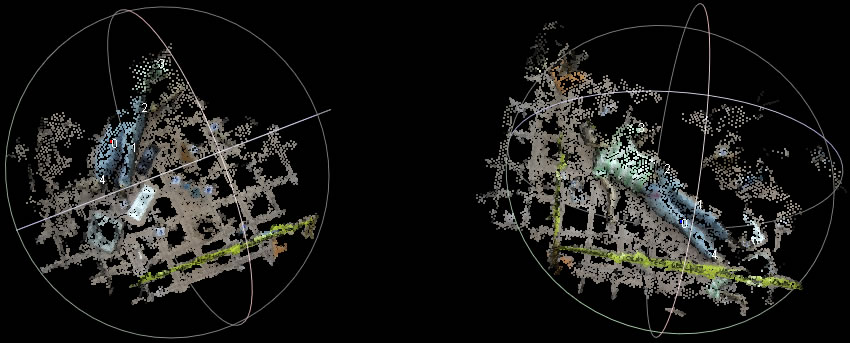}
	\end{center}
	\caption{Selection of corresponding points for \emph{ICP} algorithm.}
	\label{fig:alignment}
\end{figure}

This can be fixed using the \emph{ICP} alignment algorithm by selecting at least four points as demonstrated in Fig.~\ref{fig:alignment}.

Once a rough transformation is computed, the global alignment process starts and finds the final transformation iteratively reducing the error. For the reconstructed crime scene, this error bound was found to be 0.0010mm. The final point cloud is shown in Fig.~\ref{fig:final} below.

\begin{figure} [h!t!p]
	\begin{center}
		\includegraphics[width=0.8\columnwidth]{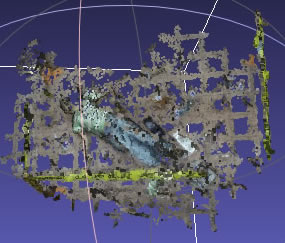}
	\end{center}
	\caption{Reconstructed crime scene in form of a 3D point cloud.}
	\label{fig:final}
\end{figure}

\section{Design Considerations}
\label{sec:design}
Having presented the details of the 3D reconstruction algorithm which uses video frames for generating the 3D model of the crime scene, this section is dedicated to define and describe design considerations and software requirements for developing a software tool for performing detailed 3D analysis on the crime scene. Key features of such a software tool can be listed as follows:

\begin{itemize}
	\item The software tool should include detailed point cloud operations involving statistical noise removal from the generated point clouds and allow various types of transformations as well as providing several view-ports.
	\item Generation of a mesh should be available from the reconstructed point cloud. The mesh should be wrapped with the texture created from images of the environment.
	\item The tool should facilitate measurement operations to calculate 3D point-to-point and mesh-to-mesh distance~\cite{Schneider2002}. Calculated distances should be annotated in the tool.
	\item Automated recognition of the items in the environment would be a very nice feature to add; however, this requires very sophisticated recognition and learning algorithms~\cite{Dang2014}.
	\item Custom annotations and comments by experts should be displayed by the system.
	\item An optional, yet nice to have feature would be creating an interface to virtual/augmented reality display~\cite{Streefkerk2013, Bostanci2013a}. This would enable and maximize the power of 3D reconstruction and present an in-situ feeling for the investigators.
	\item Finally, the tool should produce a very detailed report in a human-readable and printable format including list of items in the scene, comments by the experts and inter-item distances for archival purposes.	
\end{itemize}

\section{Conclusion}
\label{sec:conclusion}

This paper presented an automated approach for reconstructing a crime scene in 3D in order to better visualize the crime scene facilitating the investigation process. The proposed approach extracted keyframes from a video captured in the crime scene, and these keyframes were used to create sets of 3D point clouds corresponding to different parts of the scene. Finally, the crime scene was obtained in form of a single 3D point cloud after all parts are aligned together. Results show that an accurate 3D model of the environment can be easily obtained using the approach presented here. The keyframe extraction technique is prone to extracting a larger number of keyframes in case of severe motion blur; however, this is due to the detector/descriptors' performance in such extreme conditions. With the reconstruction approach described, the paper also presented some key features that should be available in a complete analysis tool.

Crime scene investigation is key to solve the nature of the crimes committed and to find the identity of the criminal. The proposed approach will provide a model that will allow the investigators to better visualize the spatial relations in the crime scene and gain a better insight into the crime through an \emph{ex-situ} analysis.

One question that may be raised here is regarding the court acceptance of the reconstructed scene as an evidence. First of all, manually modelled reconstructions can be considered as evidences. This work automatically reconstructs the crime scene from real photos which are naturally accepted as evidence. Secondly, since this reconstruction is performed using camera calibration (calculated in the bundle adjustment process), the results are metric and photo realistic. Finally, the reconstruction is based on evidence which is collected by law enforcement officers;hence, they shall be accepted by the court.

Future work will investigate employing the 3D reconstructed crime scene in a more interactive environment based on the system requirements presented here.


\section*{Acknowledgement}

The views stated in the paper are those of the authors and not necessarily those of the Turkish Armed Forces or Ministry of Interior. The author would like to thank the Turkish Gendarmarie Schools Command for providing a realistic crime scene and Second Lieutenant Suvar Azik for his support in acquiring the test dataset.

\ifCLASSOPTIONcaptionsoff
\newpage
\fi

\bibliographystyle{IEEEtr} 
\bibliography{References}

\end{document}